\pdfoutput=1
\documentclass[11pt]{article}
\usepackage{naaclhlt2009}
\usepackage{times}
\makeatletter
\newcommand{\@BIBLABEL}{\@emptybiblabel}
\newcommand{\@emptybiblabel}[1]{}
\makeatother
\usepackage{latexsym}
\usepackage{epsf}
\usepackage{graphicx}
\usepackage{hyperref}
\title{A Method for Stopping Active Learning Based on Stabilizing Predictions and the Need for User-Adjustable Stopping}

\author{
  Michael Bloodgood\Thanks{ This research was conducted while the first author was a PhD student at the University of Delaware.} \\
  Human Language Technology \\
  Center of Excellence \\
  Johns Hopkins University \\
  Baltimore, MD 21211 USA \\
  {\tt bloodgood@jhu.edu} \And
  K. Vijay-Shanker \\
  Computer and Information \\
  Sciences Department \\
  University of Delaware \\
  Newark, DE 19716 USA \\
  {\tt vijay@cis.udel.edu}}

\date{}

\begin{document}
\maketitle
\begin{abstract}
  A survey of existing methods for stopping active learning (AL) reveals the needs for methods that are: more widely applicable; more aggressive in
  saving annotations; and more stable across changing datasets. A new method for stopping AL based on stabilizing
  predictions is presented that addresses these needs. Furthermore, stopping methods are required to handle a
  broad range of different annotation/performance tradeoff valuations. Despite this, the existing body of work is dominated by
  conservative methods with little (if any) attention paid to providing users with control over the behavior of stopping methods. 
  The proposed method is shown to fill a gap in the level of aggressiveness
  available for stopping AL and supports providing users with control over stopping behavior.
\end{abstract}

\section{Introduction} \label{intro}

The use of Active Learning (AL) to reduce NLP annotation costs has generated considerable interest recently (e.g. 
\cite{bloodgood2009a,baldridge2008,zhu2008}). To realize the savings in annotation efforts that AL enables, we
must have a mechanism for knowing when to stop the annotation process. 

\begin{figure}
\begin{center}
\includegraphics[width=8cm,height=5.9cm]{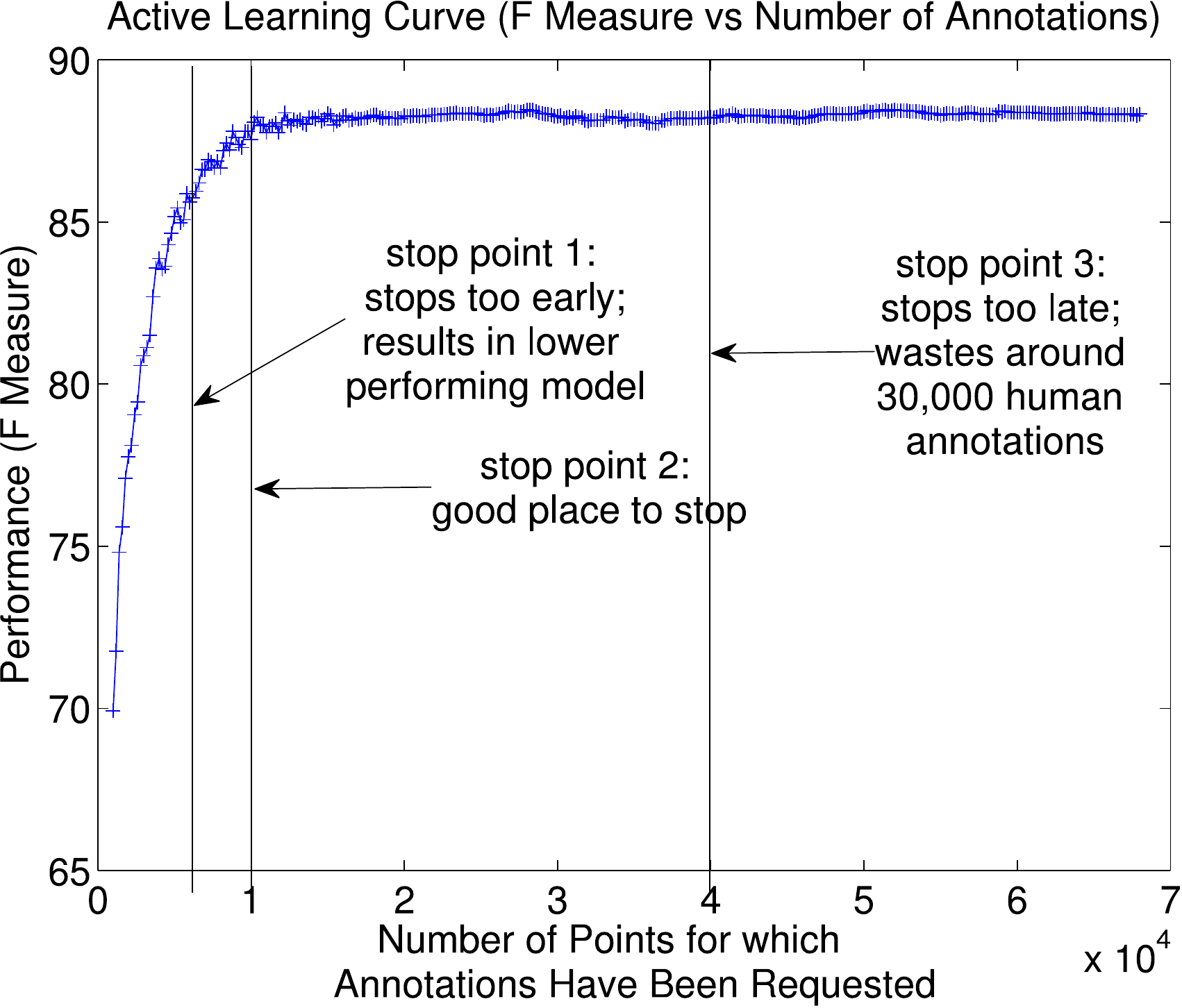}
\vspace*{-.6cm}
\caption{\label{f:motivation}Hypothetical Active Learning Curve with hypothetical stopping points.}
\end{center}
\vspace*{-.6cm}
\end{figure}

Figure~\ref{f:motivation} is intended to motivate the value of stopping at the right time. The x-axis measures the
number of human annotations that have been requested and ranges from 0 to 70,000. The y-axis measures performance in terms of
F-Measure. As can be
seen from the figure, the issue is that if we stop too early while useful generalizations are still being made, we
wind up with a lower performing system but if we stop too late after all the useful generalizations have been 
made, we just wind up wasting human annotation effort. 

The terms {\em aggressive} and {\em conservative} will be used throughout the rest of this paper to describe the behavior of 
stopping methods. Conservative methods tend to stop further to the right in Figure~\ref{f:motivation}. They are conservative 
in the sense that they're very careful not to risk losing significant amounts of F-measure, even if it means annotating many 
more examples than necessary. Aggressive methods, on the other hand, tend to stop further to the left in Figure~\ref{f:motivation}. They 
are aggressively trying to reduce unnecessary annotations. 

There has been a flurry of recent work tackling the problem of automatically determining when to stop AL (see Section~\ref{related}). 
There are three areas
where this body of work can be improved:
\begin{description}
\item[applicability] Several of the leading methods are restricted to only being used in certain situations, e.g., they 
can't be used with some base learners, they have to select points in certain batch sizes during AL, etc. (See Section~\ref{related} for 
discussion of the exact applicability constraints of existing methods.)
\item[lack of aggressive stopping] The leading methods tend to find stop points that are too far to the 
right in Figure~\ref{f:motivation}. (See Section~\ref{eval} for empirical confirmation of this.) 
\item[instability] Some of the leading methods work well on some datasets but then can completely break down on
other datasets, either stopping way too late and wasting enormous amounts of annotation effort or stopping way too early and 
losing large amounts of F-measure. (See Section~\ref{eval} for empirical confirmation of this.)
\end{description}

This paper presents a new stopping method based on stabilizing predictions that addresses each of these areas and provides user-adjustable stopping behavior.
The essential idea behind the new method is to test the predictions of the recently learned models (during AL) on examples which don't have 
to be labeled and stop when the predictions have stabilized. Some of the main advantages of the new method are that: 
it requires no additional labeled data, it's widely applicable, it fills a need for a method 
which can aggressively save annotations, it has stable performance, and it provides users with control over how aggressively/conservatively to 
stop AL. 

Section~\ref{related} discusses related work.
Section~\ref{staticPreds} explains our Stabilizing Predictions (SP) stopping criterion in detail.
Section~\ref{eval} evaluates the SP method and discusses results. 
Section~\ref{conclusions} concludes.

\section{Related Work} \label{related}

\newcite{laws2008} present stopping criteria based on the gradient of performance estimates and the gradient of
confidence estimates. Their technique with gradient of performance estimates is only applicable when probabilistic
base learners are used. The gradient of confidence estimates method is more generally applicable (e.g., it can be applied with
our experiments where we use SVMs as the base learner). This method, denoted by LS2008 in Tables and Figures, measures the rate of change of model confidence 
over a window of recent points and when the gradient falls below a threshold, AL is stopped. 

The margin exhaustion stopping criterion was developed for AL with SVMs (AL-SVM). It says to stop when all 
of the remaining unlabeled
examples are outside of the current model's margin \cite{schohn2000} and is denoted as SC2000 in Tables and
Figures. \newcite{ertekin2007b} developed a similar technique that stops when the number of support vectors saturates. 
This is equivalent to margin exhaustion in all of our experiments so this method is 
not shown explicitly in Tables and Figures.
Since we use AL with SVMs, we will compare
with margin exhaustion in our evaluation section. Unlike our SP method, margin exhaustion is only
applicable for use with margin-based methods such as SVMs and can't be used with other base learners such as Maximum Entropy, 
Naive Bayes, and
others. \newcite{schohn2000} show in their experiments that
margin exhaustion has a tendency to stop late. This is further confirmed in our experiments in Section~\ref{eval}.

The confidence-based stopping criterion
(hereafter, V2008) in \cite{vlachos2008} says to stop when model confidence consistently drops. 
As pointed out by \cite{vlachos2008}, this stopping criterion is based on the assumption that the learner/feature 
representation is
incapable of fully explaining all the examples. However, this assumption is often violated and then the 
performance of the method suffers (see Section~\ref{eval}).

Two stopping criteria (max-conf and min-err) are reported in \cite{zhu2007}. The max-conf method indicates to stop when 
the confidence of the model on each unlabeled example exceeds a threshold. In the context of margin-based methods, 
max-conf boils down to be simply a generalization of the margin exhaustion method. Min-err, reported 
to be superior to max-conf, says to stop when the accuracy of the most recent model on the current batch 
of queried examples 
exceeds some threshold (they use
0.9). \newcite{zhu2008b} proposes the use of multi-criteria-based stopping to handle setting the threshold for
min-err. They refuse to stop and they raise the min-err threshold if there have been any classification changes on
the remaining unlabeled data by consecutive actively learned models when the current min-err threshold is satisfied.
We denote this multi-criteria-based strategy, reported to work better than min-err in isolation,
by ZWH2008.
As seen in \cite{zhu2008}, sometimes
min-err indeed stops later than desired and ZWH2008 must (by nature of how it operates) stop at least as late as
min-err does. The susceptibility of ZWH2008 to stopping late is further shown emprically in Section~\ref{eval}. 
Also, ZWH2008 is not applicable for use with AL setups that select examples in small batches. 

\section{A Method for Stopping Active Learning Based on Stabilizing Predictions} \label{staticPreds}

To stop active learning at the point when annotations stop providing increases in performance,  
perhaps the most straightforward way is to use a separate set of labeled data and stop when performance 
begins to level off on that set. 
But the problem with this is that it requires additional labeled data which is counter to our original reason for using AL 
in the first place. 
Our hypothesis is that we can sense when to stop AL by looking at (only) the {\em predictions} of consecutively learned models on
examples that don't have to be labeled. 
We won't know if the predictions are correct or not but we can see if they have stabilized. 
If the predictions have stabilized, we hypothesize that the performance of the models will have stabilized 
{\em and vice-versa}, which will ensure a (much-needed) aggressive approach to saving annotations. 

SP checks for stabilization of predictions on a set of examples, called the stop set, that don't have to be labeled. 
Since stabilizing predictions on the stop set is going to be used as an indication that model stabilization has occurred, the 
stop set ought to be representative of the types of examples that will be encountered at application time. 
There are two conflicting factors in deciding upon the size of the stop set to use. On the one hand, a small set is desirable because then SP can be 
checked quickly. On the other hand, a large set is desired to ensure we don't make a decision based on a set that isn't representative of the application space. 
As a compromise between these factors, we chose a size of 2000. In Section~\ref{eval}, sensitivity analysis to stop set size is performed and more principled methods 
for determining stop set size and makeup are discussed.

It's important to allow the examples in the stop set to be queried if the active learner selects them because 
they may be highly informative and
ruling them out could hurt performance. In preliminary experiments we had made the stop set distinct from the set
of unlabeled points made available for querying and we saw performance was {\em qualitatively} the same but the AL curve 
was translated down by a few F-measure points. 
Therefore, we allow the points in the stop set to be selected during AL.\footnote{They remain in the stop set if they're
selected.} 

The essential idea is to compare successive models' predictions on the stop set to see if they have stabilized. 
A simple way to define agreement
between two models would be to measure the
percentage of points on which the models make the same predictions. 
However, experimental results on a separate development dataset show then that the cutoff agreement at which to stop is sensitive to the dataset 
being used. 
This is because different datasets have different levels of agreement that can be expected by chance and simple 
percent agreement doesn't adjust for this. 

Measurement of agreement between human annotators has received significant attention and in that context, the drawbacks of using 
percent agreement have been recognized \cite{artstein2008}. Alternative metrics have been proposed that take chance agreement into account.
In \cite{artstein2008}, a survey of several agreement metrics is presented. Most of the agreement metrics are of the form:
\begin{equation} \label{genericAgreement}
agreement = \frac{A_o - A_e}{1 - A_e},
\end{equation}
where $A_o =$ observed agreement, and $A_e =$ agreement expected by chance.
The different metrics differ in how they compute $A_e$. 

The Kappa statistic \cite{cohen1960} measures agreement expected by chance by modeling each coder (in our case model) with a separate distribution governing their
likelihood of assigning a particular category. Formally, Kappa is defined by Equation~\ref{genericAgreement} with $A_e$ computed as follows:
\begin{equation}
A_e = \sum_{k \in \{+1,-1\}} P(k|c_1) \cdot P(k|c_2),
\end{equation}
where each $c_i$ is one of the coders (in our case, models), and $P(k|c_i)$ is the probability that coder (model) $c_i$ labels an instance 
as being in category $k$. Kappa estimates $P(k|c_i)$ based on the proportion of observed instances that coder (model) $c_i$ labeled as being 
in category $k$. 

We have found Kappa to be a robust parameter that doesn't require tuning when
moving to a new dataset. On
a separate development dataset, a Kappa cutoff of 0.99 worked well. 
All of the experiments (except those in Table~\ref{t:intensity}) in the current paper 
used an agreement cutoff of Kappa = 0.99 with zero 
tuning performed. We will see in Section~\ref{eval} that this cutoff delivers robust results  
across all of the folds for all of the datasets. 

The Kappa cutoff captures the {\em intensity} of the agreement that must occur before SP will conclude to stop. 
Though an intensity cutoff of K=0.99 is an excellent default (as seen by the results in Section~\ref{eval}), 
one of the advantages of the SP method is that by giving users the option to vary the  
intensity cutoff, users can control how aggressive SP will behave. This is explored further in Section~\ref{eval}.

Another way to give users control over stopping behavior is to give them control over the {\em longevity} for which agreement (at the specified intensity) 
must be maintained before SP concludes to stop. 
The simplest implementation would be to 
check the most recent model with the previous
model and stop if their agreement exceeds the intensity cutoff.
However, independent of wanting to provide users with a longevity control, this is not an ideal approach because there's a risk that these two 
models could happen to highly agree
but then the next model will not highly agree
with them. Therefore, we propose using the average of the
agreements from a window of the k most recent
pairs of models. If we call the most recent model $M_n$, the previous model
$M_{n-1}$ and so on, with a window size of 3, we average the
agreements between $M_n$ and $M_{n-1}$, between $M_{n-1}$ and $M_{n-2}$, and between $M_{n-2}$
and $M_{n-3}$. On separate development data a window size of k=3 worked well. 
All of the experiments (except those in Table~\ref{t:longevity}) in the current paper 
used a longevity window size of k=3 with zero tuning performed. We will see in Section~\ref{eval} that this longevity default delivers robust results  
across all of the folds for all of the datasets. 
Furthermore, Section~\ref{eval} shows that varying the longevity requirement provides users with another lever for controlling how aggressively SP will behave. 

\section{Evaluation and Discussion} \label{eval}

\subsection{Experimental Setup} \label{setup}

We evaluate the Stabilizing Predictions (SP) stopping method on multiple datasets for Text Classification (TC) and Named Entity 
Recognition (NER) tasks. 
All of the datasets are freely and publicly available and have been used in many past works.

For Text Classification, we use two publicly available spam corpora: the spamassassin corpus used in
\cite{sculley2007} and the TREC spam corpus trec05p-1/ham25 described in \cite{cormack2005}. For both of these 
corpora, the task is a binary classification task and we perform 10-fold cross validation. 
We also use the Reuters dataset, in particular the Reuters-21578 Distribution 1.0 ModApte 
split\footnote{http://www.daviddlewis.com/resources/\\testcollections/reuters21578}. Since a document may belong to more than one 
category, each category is treated as a separate binary classification problem, as in \cite{joachims1998,dumais1998}. 
Consistent with \cite{joachims1998,dumais1998}, results are reported for the ten largest categories.
Other TC datasets we use are the 20Newsgroups\footnote{We used the ``bydate'' version of the dataset downloaded from
http://people.csail.mit.edu/jrennie/20Newsgroups/. This version is recommended since it makes cross-experiment 
comparison easier since there is no randomness in the selection of train/test splits.} newsgroup article classification and the WebKB web page classification datasets. 
For WebKB, as in \cite{mccallum1998b,zhu2008,zhu2008b} we use the four largest categories. 
For all of our TC datasets, we use binary features for every word that occurs in the training data at least three times. 

For NER, we use the publicly available GENIA
corpus\footnote{http://www-tsujii.is.s.u-tokyo.ac.jp/GENIA/home/\\wiki.cgi?page=GENIA+Project}.
Our features, based on those from \cite{lee2004}, are surface features such as the words in the 
named entity and two words on each side, suffix information, and positional information. 
We assume a two-phase model where boundary identification has already been performed, as in \cite{lee2004}.  

SVMs deliver high performance for the datasets we use so we employ SVMs as our base learner in the bulk of our experiments 
(maximum entropy models are used in Subsection~\ref{additional}). 
For selection of
points to query, we use the approach that was used in \cite{tong2002,schohn2000,campbell2000} of 
selecting the points that are closest to the current hyperplane. We use SVM$^{light}$ \cite{joachims1999} for training the SVMs.
For the smaller datasets (less than 50,000 examples in total), a batch size of 20 was used with an initial training set of size 100 and 
for the larger datasets (greater than 50,000 examples in total), a batch size of 200 was used with an initial training set of size 1000.
 
\subsection{Main Results} \label{results}

Table~\ref{t:stoppingAL-SVM} shows the results for all of our datasets. For each dataset, we report the 
average number of annotations\footnote{Better evaluation metrics would use more 
refined measures of annotation effort than the number of annotations because not 
all annotations require the
same amount of effort to annotate but lacking such a refined model for our datasets, we use number of annotations in these experiments.} requested by each of the stopping methods as well as the 
average F-measure achieved by each of the stopping methods.\footnote{Tests of statistical significance are performed 
using matched pairs t tests at a 95\% confidence level.} 

\begin{table*}[t]
\begin{center}
\begin{tabular}{|l|c|c|c|c|c|c|} \hline 
Task-Dataset    & SP    & V2008\protect\footnotemark & SC2000 & ZWH2008\protect\footnotemark & LS2008\protect\footnotemark & All   \\ \hline
TREC-SPAM	& 2100  & \bf 56000 & \bf 3900  & \bf 29220 & \bf 3160  & 56000 \\ \cline{2-7}
(10-fold AVG)	& 98.33 & 98.47     & 98.41     & 98.44     & 96.63     & 98.47 \\ \hline
20Newsgroups    & 678   & \bf 181   & \bf 1984  & 1340      & \bf 1669  & 11280 \\ \cline{2-7}
(20-cat AVG)    & 60.85 & \bf 18.06 & \bf 55.43 & 60.72     & \bf 54.79 & 54.81 \\ \hline
Spamassassin	& 326   & \bf 4362  & \bf 862   & \bf 398   & \bf 1176  & 5400  \\ \cline{2-7}
(10-fold AVG)	& 94.57 & 95.00     & 95.53     & 95.94     & 95.62     & 95.63 \\ \hline
NER-protein     & 8720  & \bf 67220 & \bf 17680 & \bf 18580 & \bf 2360  & 67220 \\ \cline{2-7}
(10-fold AVG)   & 89.48 & \bf 90.28 & \bf 90.38 & \bf 90.31 & \bf 76.47 & 90.28 \\ \hline
NER-DNA         & 4020  & \bf 67220 & \bf 10640 & \bf 7200  & \bf 3900  & 67220 \\ \cline{2-7}
(10-fold AVG)   & 82.40 & \bf 84.31 & \bf 84.73 & \bf 84.51 & \bf 74.74 & 84.31 \\ \hline
NER-cellType    & 3840  & \bf 29600 & \bf 5540  & 11580     & 4580      & 67220 \\ \cline{2-7}
(10-fold AVG)   & 86.15 & 86.87     & \bf 87.19 & \bf 87.32 & 85.65     & 87.83 \\ \hline
Reuters         & 484   & \bf 6762  & \bf 1196  & \bf 650   & \bf 1272  & 9580 \\ \cline{2-7}
(10-cat AVG)    & 74.29 & \bf 65.81 & \bf 73.88 & 76.77     & 74.00     & 75.64 \\ \hline
WKB-Course      & 790   & \bf 184   & \bf 1752  & \bf 912   & \bf 1740  & 7420 \\ \cline{2-7}
(10-fold AVG)   & 83.12 & \bf 30.34 & \bf 80.47 & 83.16     & \bf 80.55 & 80.19 \\ \hline
WKB-Faculty     & 808   & 892       & \bf 1932  & \bf 1062  & \bf 1818  & 7420 \\ \cline{2-7}
(10-fold AVG)   & 81.53 & \bf 40.14 & 81.79     & 81.64     & 81.99     & 82.36 \\ \hline
WKB-Project     & 646   & 916       & \bf 1358  & \bf 794   & \bf 1482  & 7420 \\ \cline{2-7}
(10-fold AVG)   & 63.30 & \bf 25.33 & \bf 58.11 & 61.82     & \bf 59.30 & 61.19 \\ \hline
WKB-Student     & 1258  & 894       & \bf 2400  & \bf 1468  & \bf 2150  & 7420 \\ \cline{2-7}
(10-fold AVG)   & 84.70 & \bf 50.66 & \bf 83.46 & 84.39     & \bf 83.19 & 83.30 \\ \hline
Average         & 2152  & \bf 21294 & \bf 4477  & \bf 6655  & 2301      & 28509 \\ \cline{2-7}
(macro-avg)     & 81.70 & \bf 62.30 & 80.85     & 82.27     & \bf 78.45	& 81.27 \\ \hline										 								    		    	     
\end{tabular}
\end{center}
\caption{\label{t:stoppingAL-SVM} Methods for stopping AL. For each dataset, the average number of annotations at the
automatically determined stopping points and the average F-measure at the automatically 
determined stopping points are displayed. \bf Bold entries \rm are statistically significantly different than
SP (and non-bold entries are not). The Average row is simply an unweighted macro-average over all the
datasets. The final column (labeled "All") represents standard fully supervised passive learning with the entire set of training data.}
\end{table*}
\addtocounter{footnote}{-2}\footnotetext{\cite{vlachos2008} 
suggests using three drops in a row to detect a consistent drop in confidence so we do the same in our
implementation of the method from \cite{vlachos2008}.}	
\addtocounter{footnote}{1}\footnotetext{Following \cite{zhu2008b}, we 
set the starting accuracy threshold to 0.9 when reimplementing their method.} 
\addtocounter{footnote}{1}\footnotetext{\cite{laws2008} uses a window of size 100 and a threshold of 0.00005 so 
we do the same in our re-implementation of their method.} 

There are two facts worth keeping in mind. First, the numbers in Table~\ref{t:stoppingAL-SVM} are 
averages and therefore, sometimes two methods could have very similar average numbers of annotations but wildly different average F-measures 
(because one of the methods was consistently stopping around its average whereas the other was stopping way too early and way too late). 
Second, sometimes a method with a higher average number of annotations has a lower average F-measure than a method with a lower average number of annotations.
This can be caused because of the first fact just mentioned about the numbers being averages and/or this can also be caused by the "less is more" phenomenon 
in active learning where often with less data, a higher-performing model is learned than with all the data; this was first reported in \cite{schohn2000} and 
subsequently observed by many others (e.g., \cite{vlachos2008,laws2008}). 

There are a few observations to highlight regarding the performance of the various stopping methods:
\begin{itemize}
\item SP is the most parsimonious method in terms of annotations. It stops the earliest and remarkably it is able
to do so largely without sacrificing F-measure. 
\item All the methods except for SP and SC2000 are unstable in the sense that on at least one dataset they have a
major failure, either stopping way too late and wasting large numbers of annotations (e.g. ZWH2008 and V2008 on TREC Spam)
or stopping way too early and losing large amounts of F-measure (e.g. LS2008 on NER-Protein)  .
\item It's not always clear how to evaluate stopping methods because the tradeoff between the value of extra F-measure 
versus saving annotations is not clearly known and will be different for different applications and users. 
\end{itemize}
This last point deserves some more discussion. In some cases it is clear that one stopping method is the best. For
example, on WKB-Project, the SP method saves the most annotations {\em and} has the highest F-measure. But which method
performs the best on NER-DNA? Arguments can reasonably be made for SP, SC2000, or ZWH2008 being the best in
this case depending on what exactly the annotation/performance tradeoff is. A promising direction for 
research on AL stopping methods is to develop user-adjustable stopping methods that stop as
aggressively as the user's annotation/performance preferences dictate. 

One avenue of providing user-adjustable stopping is that if some methods are known to perform consistently in an
aggressive manner against annotating too much while others are known to perform consistently in a conservative
manner, then users can pick the stopping criterion that's more suitable for their particular 
annotation/performance valuation.
For this purpose, SP fills a gap as the other stopping criteria seem to be conservative in the sense defined in Section~\ref{intro}. 
SP, on the other hand, is more of an
aggressive stopping criterion and is less likely to annotate data that is not needed. 

A second avenue for providing user-adjustable stopping is a single stopping method that is itself adjustable. 
To this end, Section~\ref{additional} shows how {\em intensity} and {\em longevity} provide levers that can be used 
to control the behavior of SP in a controlled fashion. 

\begin{figure}
\begin{center}
\includegraphics[width=8cm,height=6.2cm]{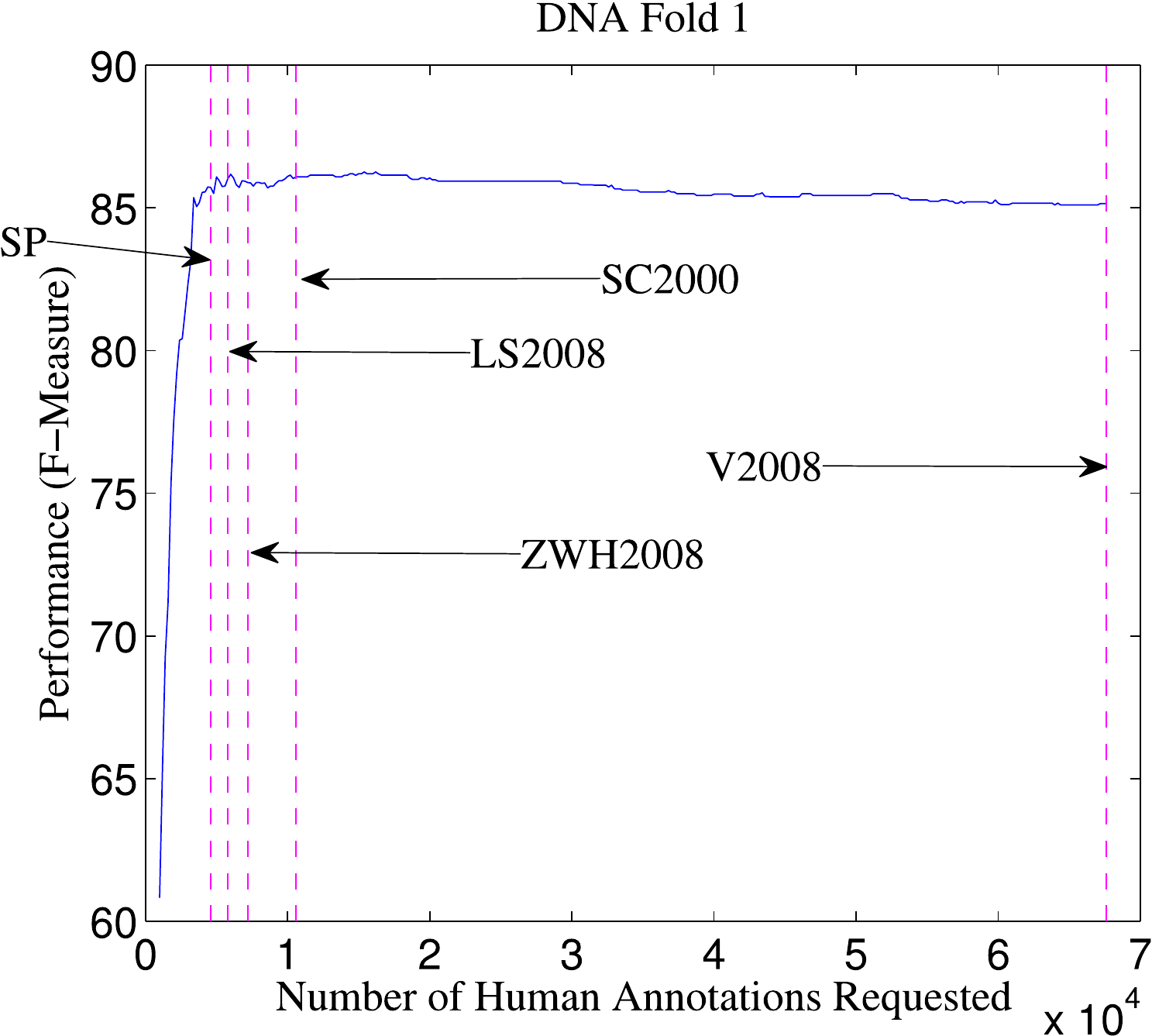}
\caption{\label{f:dna1}Graphic with stopping criteria in action for fold 1 of NER of DNA from the GENIA corpus. The x-axis ranges
from 0 to 70,000.}
\end{center}
\end{figure}

Sometimes viewing the stopping points of the various criteria on a graph with the active learning curve can help
one visualize how the methods perform. 
Figure~\ref{f:dna1} shows the graph for a representative
fold.\footnote{It doesn't make sense to show a graph for the
average over cross validation because the average number of annotations at the stopping point may cross the learning curve at a
completely misleading point. Consider a method that stops way too early and way too late at times.} The x-axis measures the number of human annotations that have
been requested so far. The y-axis measures performance in terms of F-Measure. The vertical lines are where the various
stopping methods would have stopped AL if we hadn't continued the simulation. 
The figure reinforces and illustrates what we have seen 
in Table~\ref{t:stoppingAL-SVM}, namely that SP stops
more aggressively than existing criteria and is able to do so without sacrificing performance. 

\subsection{Additional Experiments} \label{additional}

All of the additional experiments in this subsection were conducted on our least computationally demanding dataset, Spamassassin.
The results in Tables~\ref{t:intensity} and \ref{t:longevity} show how varying the intensity cutoff and 
the longevity requirement, respectively, of SP enable a user to control stopping behavior.
Both methods enable a user to adjust stopping in a controlled fashion (without radical changes in behavior). 
Areas of future work include:
combining the intensity and longevity methods for controlling behavior; and developing precise expectations on the change in behavior
corresponding to changes in the intensity and longevity settings.

\begin{table}
\begin{center}
\begin{tabular}{|l|c|c|} \hline
Intensity & Annotations & F-Measure \\ \hline
K=99.5    & 364         & 96.01 \\ \hline
K=99.0    & 326         & 94.57 \\ \hline
K=98.5    & 304         & 95.59 \\ \hline
K=98.0    & 262         & 93.75 \\ \hline
K=97.5    & 242         & 93.35 \\ \hline
K=97.0    & 224         & 90.91 \\ \hline
\end{tabular}
\end{center}
\caption{\label{t:intensity} Controlling the behavior of stopping through the use of {\em intensity}. For Kappa intensity levels in \{97.0, 97.5, 98.0, 98.5, 99.0, 99.5\}, the 10-fold 
average number of annotations at the
automatically determined stopping points and the 10-fold average F-measure at the automatically 
determined stopping points are displayed for the Spamassassin dataset.}
\end{table}	

\begin{table}[t]
\begin{center}
\begin{tabular}{|l|c|c|} \hline
Longevity & Annotations & F-Measure \\ \hline
k=1       & 284         & 95.17 \\ \hline
k=2       & 318         & 94.95 \\ \hline
k=3       & 326         & 94.57 \\ \hline
k=4       & 336         & 95.40 \\ \hline
k=5       & 346         & 96.41 \\ \hline
k=6       & 366         & 94.53 \\ \hline
\end{tabular}
\end{center}
\caption{\label{t:longevity} Controlling the behavior of stopping through the use of {\em longevity}. For window length k longevity levels in \{1, 2, 3, 4, 5, 6\}, the 10-fold 
average number of annotations at the
automatically determined stopping points and the 10-fold average F-measure at the automatically 
determined stopping points are displayed for the Spamassassin dataset.}
\end{table}	

\begin{table}
\begin{center}
\begin{tabular}{|l|c|c|} \hline
Stop Set Size & Annotations & F-Measure \\ \hline
2500          & 326         & 95.58 \\ \hline
2000          & 326         & 94.57 \\ \hline
1500          & 314         & 95.00 \\ \hline
1000          & 328         & 95.73 \\ \hline
500           & 314         & 94.57 \\ \hline
\end{tabular}
\end{center}
\caption{\label{t:stopSizes} Investigating the sensitivity to stop set size. For stop set sizes in \{2500, 2000, 1500, 1000, 500\}, the 10-fold 
average number of annotations at the
automatically determined stopping points and the 10-fold average F-measure at the automatically 
determined stopping points are displayed for the Spamassassin dataset.}
\end{table}

\begin{table*}[t]
\begin{center}
\begin{tabular}{|l|c|c|c|c|c|} \hline 
Task-Dataset    & SP    & V2008      & ZWH2008   & LS2008    & All   \\ \hline
Spamassassin	& 286   & 1208       & \bf 386   & \bf 756   & 5400  \\ \cline{2-6}
(10-fold AVG)	& 94.92 & \bf 89.89  & 95.31     & 96.40     & 91.74 \\ \hline
\end{tabular}
\end{center}
\caption{\label{t:stoppingAL-maxEnt} Methods for stopping AL with maximum entropy as the base learner. For each stopping method, the average number of annotations at the
automatically determined stopping point and the average F-measure at the automatically 
determined stopping point are displayed. \bf Bold entries \rm are statistically significantly different than
SP (and non-bold entries are not). 
SC2000, the margin exhaustion method, is not shown since it can't be used with a non-margin-based learner.
The final column (labeled "All") represents standard fully supervised passive learning with the entire set of training data.}
\end{table*}		

The results in Table~\ref{t:stopSizes} show results for different stop set sizes. Even with random selection of a stop set as small as 500, SP's performance holds fairly steady. 
This plus the fact that random selection of stop sets of size 2000 worked across all the folds of all the datasets in Table~\ref{t:stoppingAL-SVM} show that in practice perhaps the 
simple heuristic of choosing a fairly large random set of points works well. 
Nonetheless, we think the size necessary will depend on the dataset and other factors such as the feature representation so 
more principled methods of determining the size and/or the makeup of the stop set are an area for future work. For example, construction techniques could be 
developed to create stop sets with high representativeness (in terms of feature space) 
density (meaning representativeness of stop set divided by size of stop set). 
For example, a possibility is to cluster examples before AL begins and then make sure the
stop set contains examples from each of the clusters. Another possibility is to use a greedy 
algorithm where the stop set is iteratively grown where on each iteration the center of mass of the stop set in feature space is computed 
and an example in the unlabeled pool that is maximally far in feature space from this center of mass is selected for inclusion in the stop set. 
This could be useful for efficiency (in terms of getting the same stopping performance 
with a smaller stop set as could be achieved with a larger
stop set) and also as a way to ensure adequate representation of the task space. 
The latter can be accomplished by perhaps continuing to add
examples to the stop set until adding new examples is no longer increasing the representativeness of the stop set.

\begin{figure}
\begin{center}
\includegraphics[width=8cm,height=6.2cm]{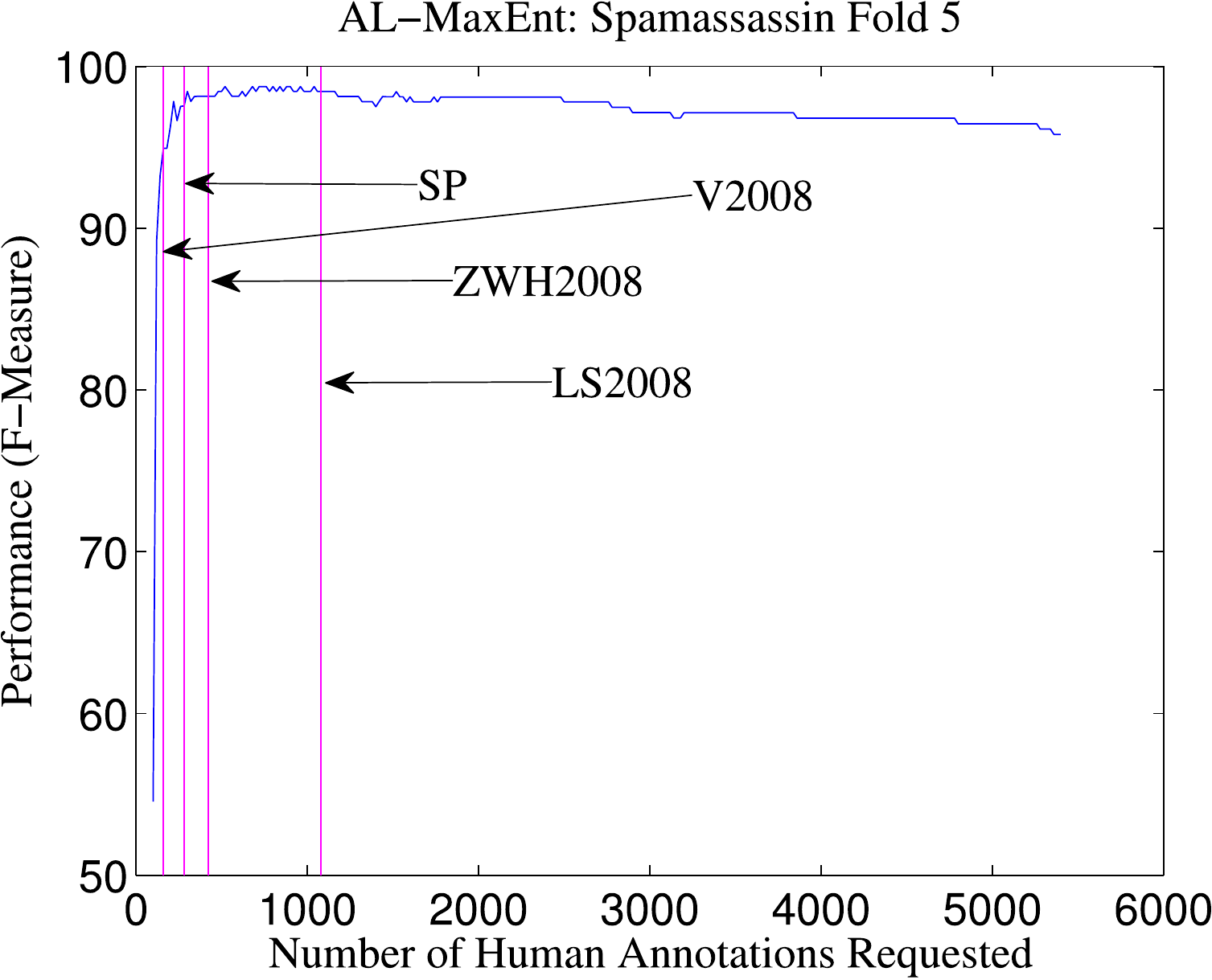}
\caption{\label{f:spam5}Graphic with stopping criteria in action for fold 5 of TC of the spamassassin corpus. The x-axis ranges
from 0 to 6,000.}
\end{center}
\end{figure}

As one of the advantages of SP is that it's widely applicable, Table~\ref{t:stoppingAL-maxEnt} shows 
the results when using maximum entropy models as the base learner during AL (the query points selected are those which the model is most uncertain about). The results reinforce our 
conclusions from the SVM experiments, with SP performing aggressively and all statistically significant differences in performance being 
in SP's favor. Figure~\ref{f:spam5} shows the graph for a representative fold.

\section{Conclusions} \label{conclusions}

Effective methods for stopping AL are crucial for realizing the potential annotation savings 
enabled by AL. A survey of existing stopping methods identified three areas where 
improvements are called for. The new stopping method based on Stabilizing Predictions (SP) addresses all three areas: 
SP is widely applicable, stable, and aggressive
in saving annotations. 

The empirical evaluation of SP and the existing methods was informative for evaluating the criteria but it was 
also informative for demonstrating the difficulties for rigorous objective evaluation of stopping criteria due to
different annotation/performance tradeoff valuations. This opens up a future area for work on user-adjustable
stopping. Two potential avenues for enabling user-adjustable stopping 
are a single criterion that is itself adjustable or a suite of methods with consistent
differing levels of aggressiveness/conservativeness from which users can pick the one(s) that suit their
annotation/performance tradeoff valuation. SP substantially widens the range of behaviors of existing methods that users 
can choose from. Also, SP's behavior itself can be adjusted through user-controllable parameters.  
 
\bibliographystyle{naaclhlt2009}
\bibliography{paper}

\end{document}